\documentclass[conference]{IEEEtran}

\IEEEoverridecommandlockouts
\usepackage{cite}
\usepackage{amsmath,amssymb,amsfonts}
\usepackage{algorithmic}
\usepackage{graphicx}
\usepackage{textcomp}
\usepackage{xcolor}
\def\BibTeX{{\rm B\kern-.05em{\sc i\kern-.025em b}\kern-.08em
    T\kern-.1667em\lower.7ex\hbox{E}\kern-.125emX}}
\begin{document}

\title{Forecasting Evolution of Clusters in Game Agents with Hebbian Learning\\
\thanks{This work was supported in part by the Office of Naval Research under Grant N00014-20-1-2432. The views and conclusions contained in this document are those of the authors and should not be interpreted as representing the official policies, either expressed or implied, of the Office of Naval Research or the U.S. Government.}
}

\author{
\IEEEauthorblockN{Beomseok Kang, Saibal Mukhopadhyay}
\IEEEauthorblockA{\textit{School of Electrical and Computer Engineering} \\
\textit{Georgia Institute of Technology}\\
Atlanta, GA, USA \\
\{beomseok, smukhopadhyay6\}@gatech.edu}
}

\maketitle

\begin{abstract}
Large multi-agent systems such as real-time strategy games are often driven by collective behavior of agents. For example, in StarCraft II, human players group spatially near agents into a team and control the team to defeat opponents. In this light, clustering the agents in the game has been used for various purposes such as the efficient control of the agents in multi-agent reinforcement learning and game analytic tools for the game users. However, despite the useful information provided by clustering, learning the dynamics of multi-agent systems at a cluster level has been rarely studied yet. In this paper, we present a hybrid AI model that couples unsupervised and self-supervised learning to forecast evolution of the clusters in StarCraft II. We develop an unsupervised Hebbian learning method in a set-to-cluster module to efficiently create a variable number of the clusters with lower inference time complexity than K-means clustering. Also, a long short-term memory based prediction module is designed to recursively forecast state vectors generated by the set-to-cluster module to define cluster configuration. We experimentally demonstrate the proposed model successfully predicts complex movement of the clusters in the game.
\end{abstract}

\begin{IEEEkeywords}
Clustering, Multiagent systems, Artificial intelligence in games, Hebbian learning, Unsupervised learning.
\end{IEEEkeywords}

\section{Introduction}
Collective behavior, such as collaboration and competition, is often observed in large multi-agent systems. While many artificial intelligence (AI) models have been developed to understand the dynamics of individual agents, they may be redundant if the systems are mainly driven by collective behavior. A good example is real-time strategy (RTS) games, such as StarCraft II, where multiple players control hundreds of game agents and build strategies to defeat opponents \cite{ontanon2013survey, berner2019dota, ye2020towards}. Given the large number of agents, players frequently group the spatially near agents as a team and determine how to arrange and control the team to survive in complex battles. In other words, the dynamics of the agents in the game are often defined by groups of agents. Therefore, a natural question in such systems is whether we can design an AI model that can efficiently learn and forecast the evolution of game agents at the cluster level, instead of predicting every individual agent as prior works have generally done \cite{kang2022unsupervised}.

\begin{figure}
\centering
\includegraphics[width=\columnwidth]{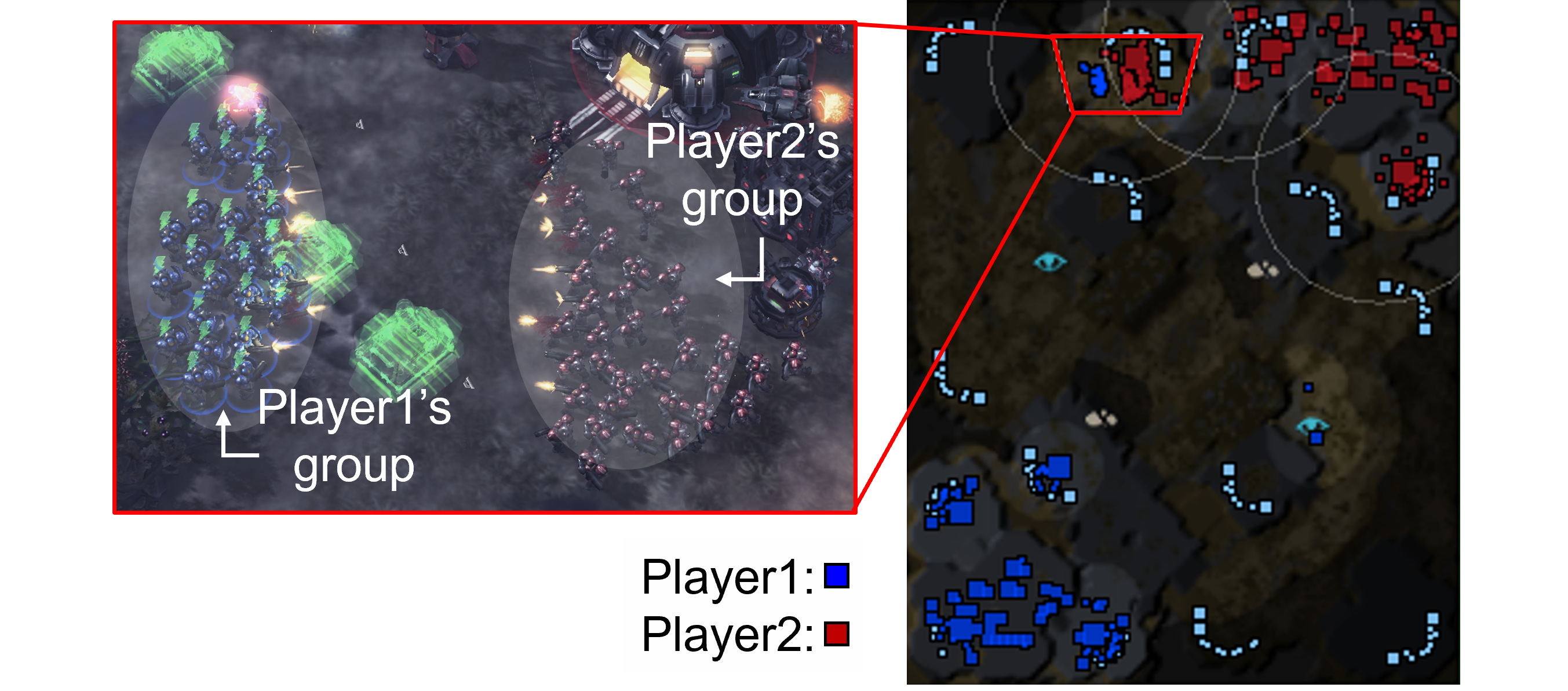}
\caption{StarCraft II game map given to players (right) and snapshot in the game where combat is observed (left). The left bottom and right top area are territories for player1 and player2. Players frequently control a large group of the game agents in combats.}
\label{figure_game_map}
\end{figure}

Clustering has been widely applied to game agents for various purposes \cite{ontanon2013survey, kuan2017visualizing, shao2018starcraft, vcerticky2018starcraft}. Fig. \ref{figure_game_map} shows a crop of the StarCraft II game map in which the agents of two players are marked with different colors. As the figure displays groups of the agents in a combat scenario, several strategies are used to be described by spatio-temporal changes in the agent clusters \cite{kuan2017visualizing}. A general strategy used in combats is to first gather soldier agents into a cluster and then kill enemies one by one instead of attacking multi enemies \cite{shao2018starcraft}. In this light, tactics in StarCraft II are closely related to dynamics of the agent clusters \cite{ontanon2013survey}, and learning dynamics of the clusters will allow the models to efficiently understand underlying tactics.

Apart from the strategic understanding, other usefulness of clustering the game units has been also reported. Recent studies on multi-agent reinforcement learning in the game have exploited clustering to assign the same action to a group of the near agents so that the large action space is reduced \cite{baek2019efficient, farquhar2020growing, vcerticky2018starcraft}. The motivation behind the clustering is that the spatially close agents are likely to perform similar actions \cite{justesen2014script}. Furthermore, clustering groups of the agents has been utilized in visualization instruments for game analytic purposes \cite{wallner2020multivariate, kuan2017visualizing}. It helps game users to easily discover the motion of many units and quickly identify important events such as a large number of dead. That is, the clusters provide the useful information such as strategy and scene understanding and efficient control policies to the external observers, which may not be restricted only to the game. However, the prior works do not aim to learn and predict the dynamics of the agent clusters, limiting the clustering applications to the currently observed status of the system.

In this paper, we propose a hybrid AI model that combines unsupervised (Hebbian) and self-supervised (gradient-based) learning to forecast the evolution of agent clusters in StarCraft II. Our model is designed with a set-to-cluster module and a prediction module. The set-to-cluster module represents a set of agents with regard to cluster centroids and radii into a state vector. The key innovation of the clustering module is to efficiently create a variable number of clusters without pre- and post- processing on the number of clusters by users, which is important as the number of clusters frequently varies in the heat of battles. We train the model using unsupervised Hebbian learning to find the optimal location of cluster centroids and apply a Winner-take-all algorithm to activate a variable number of neurons that represent the cluster centroids. The long short-term memory (LSTM) based prediction module recursively forecasts the state vectors generated by the set-to-cluster module in the next time steps, and LSTM is trained in a self-supervised setting using gradient based learning. 

Hebbian learning is a biologically inspired unsupervised algorithm that updates a synaptic weight where two adjacent neurons fire together. There have been several applications with Hebbian learning in image processing, reinforcement learning, and clustering \cite{miconi2021multi, amato2019hebbian, najarro2020meta, miconi2018differentiable, do2007growing, hu2014modeling}. However, these methods have been evaluated by simple dataset such as MNIST or dealt with a single agent \cite{krotov2019unsupervised, najarro2020meta}. More importantly, most of the prior works on Hebbian learning have not been considered in prediction models for complex and dynamic multi-agent systems. We mainly describe our Hebbian learning approach for clustering in section II. The prediction model and training procedure for it are explained there as well. Dataset and cluster prediction results are described in section III, and there the details in the performance such as failure analysis and comparison with other method are also considered. This paper makes the following key contributions:
\begin{itemize}
\item We formulate a novel task to forecast evolution of multiple agents at a cluster level in a real-time strategy game driven by collective behavior.
\item We present a hybrid AI model that couples an unsupervised (Hebbian) set-to-cluster module and self-supervised (gradient-based) prediction module. 
\item Our set-to-cluster module is able to represent the cluster configuration such as centroids and radii and efficiently create a variable number of the clusters with lower computational complexity than K-mean clustering.
\item We empirically demonstrate that the proposed model is able to forecast the complex evolution of the agent clusters such as merger and division in a StarCraft II replay video.
\end{itemize}

\section{Proposed Approach}
\subsection{Set to Cluster Module}

\begin{figure}
\centering
\includegraphics[width=\columnwidth]{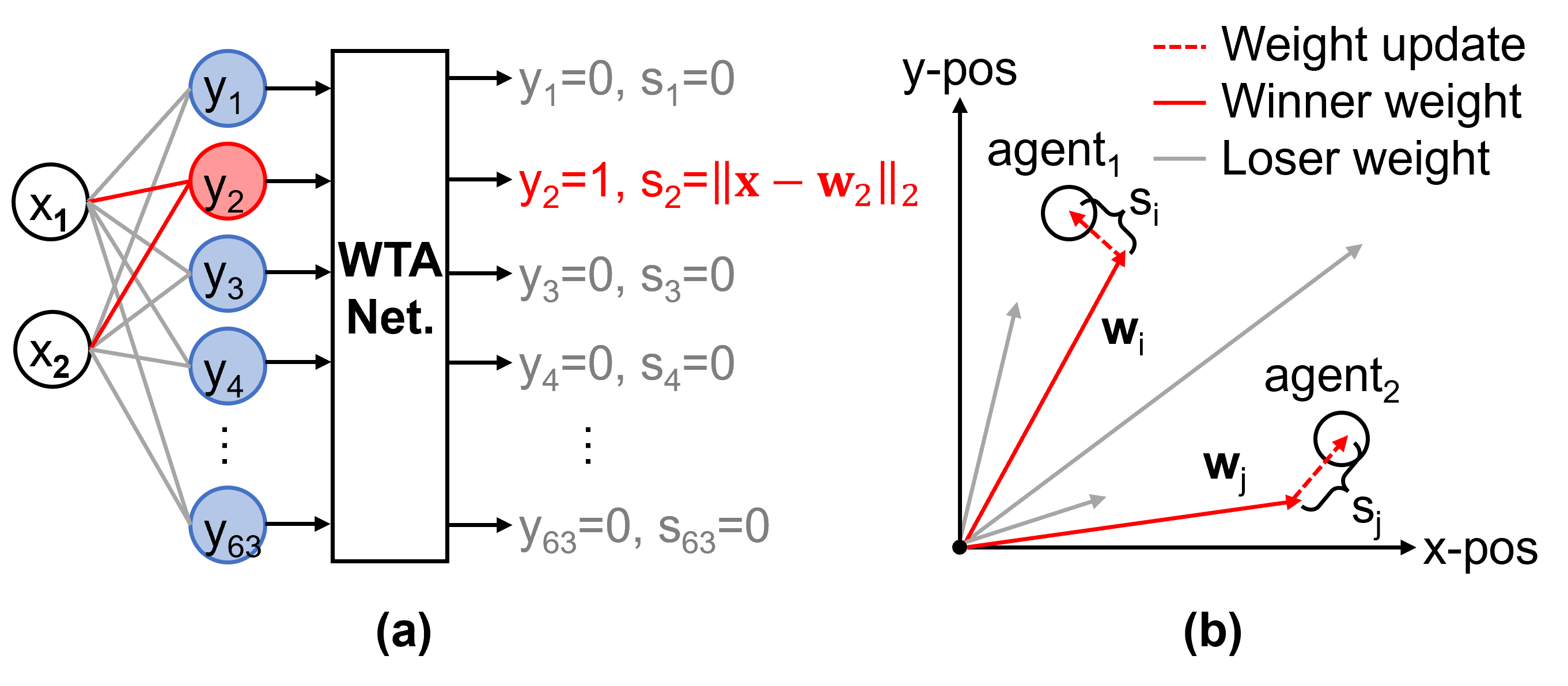}
\caption{Schematic of fully connected layer in set-to-cluster module (a) and schematic to describe weight vectors and agents in clustering (b).}
\label{figure_cluster_encoder}
\end{figure}
The input to the proposed model is a set of agent positions. Note, the number of the agents in StarCraft II is an actively changing variable. Our first step to learn dynamics of the agent clusters is to encode the position set in a frame to a vector so that LSTM is trained in fixed dimension latent space. We design the set-to-cluster module based on PointNet consisted of MLP and a MaxPooling layer \cite{qi2017pointnet}. We apply Hebbian learning on a fully-connected (FC) layer in the set-to-cluster module, and the other modules are trained by supervised learning.

\begin{figure*}
\begin{center}
\includegraphics[width=\textwidth]{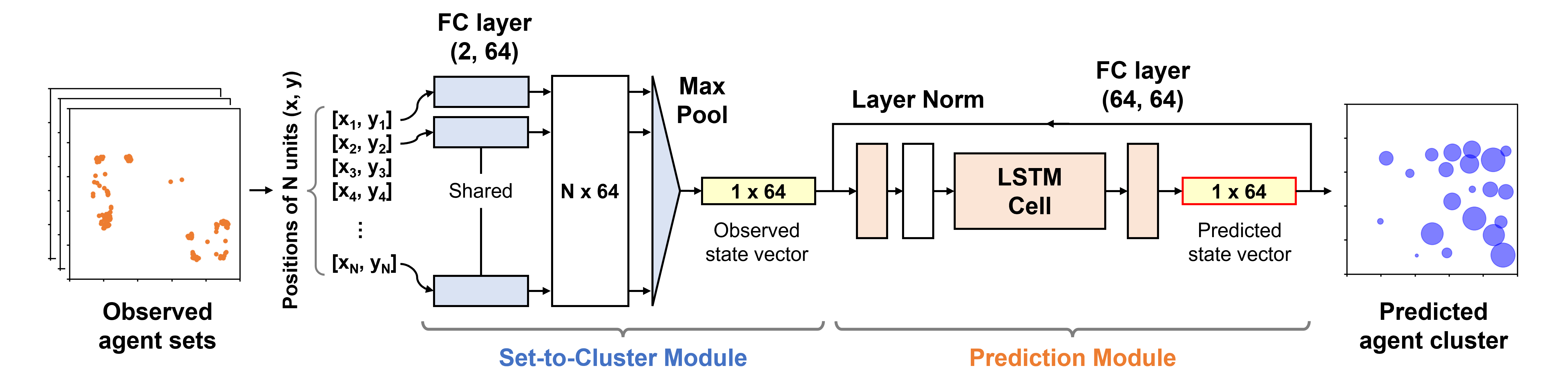}
\end{center}
\caption{Schematic of entire model design for cluster prediction. The set-to-cluster module generates state vectors from observed agent sets, and the prediction module recursively predicts state vectors in the next time steps.}
\label{figure_cluster_prediction_model}
\end{figure*}

Fig. \ref{figure_cluster_encoder}(a) shows the module design which input is a positional vector of an agent (\(x_{1}, x_{2}\)). We define output \(y\) and state \(s\); the output represents which cluster centroid is activated for an agent position, and the state indicates the Euclidean distance (\(\|\textbf{x}-\textbf{w}\|_{2}\)) between the agent and the activated cluster centroid. As shown in the figure, the output and state are one-hot vectors due to a Winner-take-all (WTA) network. It competitively activates a single neuron which weight vector is closest in terms of the Euclidean distance from the input positional vector, namely a winner neuron. For loser neurons, the output values are forced to be zero, also indicating the radii of these clusters are zero. An important idea here is that the module uses the weight vectors as cluster centroids so that the neuron activation indicates where the associated cluster centroid is by looking at its weight vector. We restrict the minimum radius of winner neurons to 1e-2 to avoid the zero radii when the agent is exactly on a cluster centroid. We differentiate an output vector (\(\textbf{y}\)) and state vector (\(\textbf{s}\)) in the sense that a cluster radius is not considered in the output vector. The output and state vectors are mathematically described as:
\begin{equation}
\label{eq2}
y_{i}=
    \begin{cases}
      1 & \text{if} \: i = \underset{j}{\mathrm{arg\:min}}\:{\|\textbf{x}-\textbf{w}_{j}\|_{2}} \\
      0 & \text{otherwise}
    \end{cases}
\end{equation}

\begin{equation}
\label{eq3}
s_{i}=
    \begin{cases}
      \text{max}(0.01, \: \|\textbf{x}-\textbf{w}_{i}\|_{2}) & \text{if} \: i = \underset{j}{\mathrm{arg\:min}}\:{\|\textbf{x}-\textbf{w}_{j}\|_{2}} \\
      0 & \text{otherwise}
    \end{cases}
\end{equation}

\noindent where \(i\) indicates the \(i\)-th element of vectors.

Now, consider the multiple input positional vectors. The same set-to-cluster module processes each positional vector and represent it into the state vector. Then, a MaxPooling layer, which is omitted in the figure, aggregates state vectors to find the maximum state value for each neuron. As some of the agents will be associated with the same winner neuron, the maximum state indicates the largest distance between the cluster centroid and associated agents, which is the radius of the cluster at the end. The number of non-zero element in the pooled state vector is the number of the clusters. From here, we simply call the pooled state vector as state vector. With the state vector and corresponding weight vectors, we can directly interpret the cluster configuration without an additional decoding process.

\subsection{Hebbian Learning for Clustering}

There are different types of Hebbian learning rules. The simplest rule defines that a weight update is proportional to the multiplication of input and output activation. A variant, Grossberg’s instar rule, additionally includes the multiplication of output and a weight vector in the basic Hebbian learning rule as in (\ref{equation_basic_rule}) \cite{grossberg1976adaptive}.
\begin{equation}
\Delta{\textbf{w}} \propto y(\textbf{x}-\textbf{w})
\label{equation_basic_rule}
\end{equation}

\noindent It updates a weight vector (\(\textbf{w}\)) to be close to an input vector (\(\textbf{x}\)) if output (\(y\)) is activated. We apply the modified Hebbian learning algorithm on the FC layer where input is a position of the agent and output is a state vector to represent which centroid is closest from the agent and the distance between them.

Hu et al. have shown that K-means clustering can be realized by Hebbian learning and Winner-take-all \cite{hu2014modeling}. However, the prior work attempts to hierarchically cluster natural images with deep belief networks, where the creation of a variable number of the clusters is not necessary. Also, the input dimension is assumed to be fixed while we need to deal with a variable number of the positional vectors. We take advantage of the basic concept in \cite{hu2014modeling}, but the key difference is in processing a set of the agents and creating a variable number of the clusters, which is enabled by the proposed clustering module with Hebbian learning.

Fig. \ref{figure_cluster_encoder}(a) shows that WTA algorithm is applied to an output vector (\(\textbf{y}\)) to competitively fire a single output neuron where \(\|\textbf{x}-\textbf{w}\|_{2}\) is the minimum. Then, we can re-write (\ref{equation_basic_rule}) to (\ref{equation_learning_rule}) by introducing a function \(f(|\textbf{x}-\textbf{w}|)\):
\begin{equation}
  f(|\textbf{x}_{i}-\textbf{w}_{j}|) =  
    \begin{cases}
      1 & \text{if} \: i = \underset{k}{\mathrm{arg\:min}}\:{|\textbf{x}_{k}-\textbf{w}_{j}|} \\
      0 & \text{otherwise}
    \end{cases}       
\label{equation_distance_training}
\end{equation}

\begin{equation}
\Delta\textbf{w}_{j} = \eta\frac{1}{N}\sum_{i}{f(|\textbf{x}_{i}-\textbf{w}_{j}|)(\textbf{x}_{i}-\textbf{w}_{j})}
\label{equation_learning_rule}
\end{equation}

\noindent where \(\eta\) is the learning rate, and \(N\) is the number of the agents being processed. The function determines whether a weight vector (\(\textbf{w}_{j}\)) is to be updated depending on the Euclidean distance from an input position (\(\textbf{x}_{i}\)). As the weight update is averaged by the N agents, the weight vector (\(\textbf{w}_{j}\)) is gradually updated to the average position of the agents associated with the \(j\)-th neuron. That is, Hebbian learning reduces the radius of the cluster to be represented during the training. After the training, the initial cluster centroids are moved to the areas where the agents frequently appeared. The set-to-cluster module uses the weight vectors as the pre-defined cluster centroids, thereby, saving computational costs for the optimal centroids used to be searched every inference in conventional unsupervised clustering methods. However, it assumes that the learned cluster centroids will still be effective for test datasets. For example, players may not be able to locate agent groups near the learned cluster centroids if the geography of the game map in test datasets is really different with training datasets.

Fig. \ref{figure_cluster_encoder}(b) describes the weight update by Hebbian learning in a simple scenario with two agents. As each agent is individually processed, two states are created with different winner neurons. For example, \(\textbf{w}_{i}\) is the weight vector of a winner neuron for the agent\(_{1}\) and \(\textbf{w}_{j}\) is for the agent\(_{2}\). The dotted vector in the figure indicates the weight updates for the winner neurons while the other weight vectors of loser neurons remain at the same positions.

\begin{figure}[t]
\begin{center}
\includegraphics[width=\columnwidth]{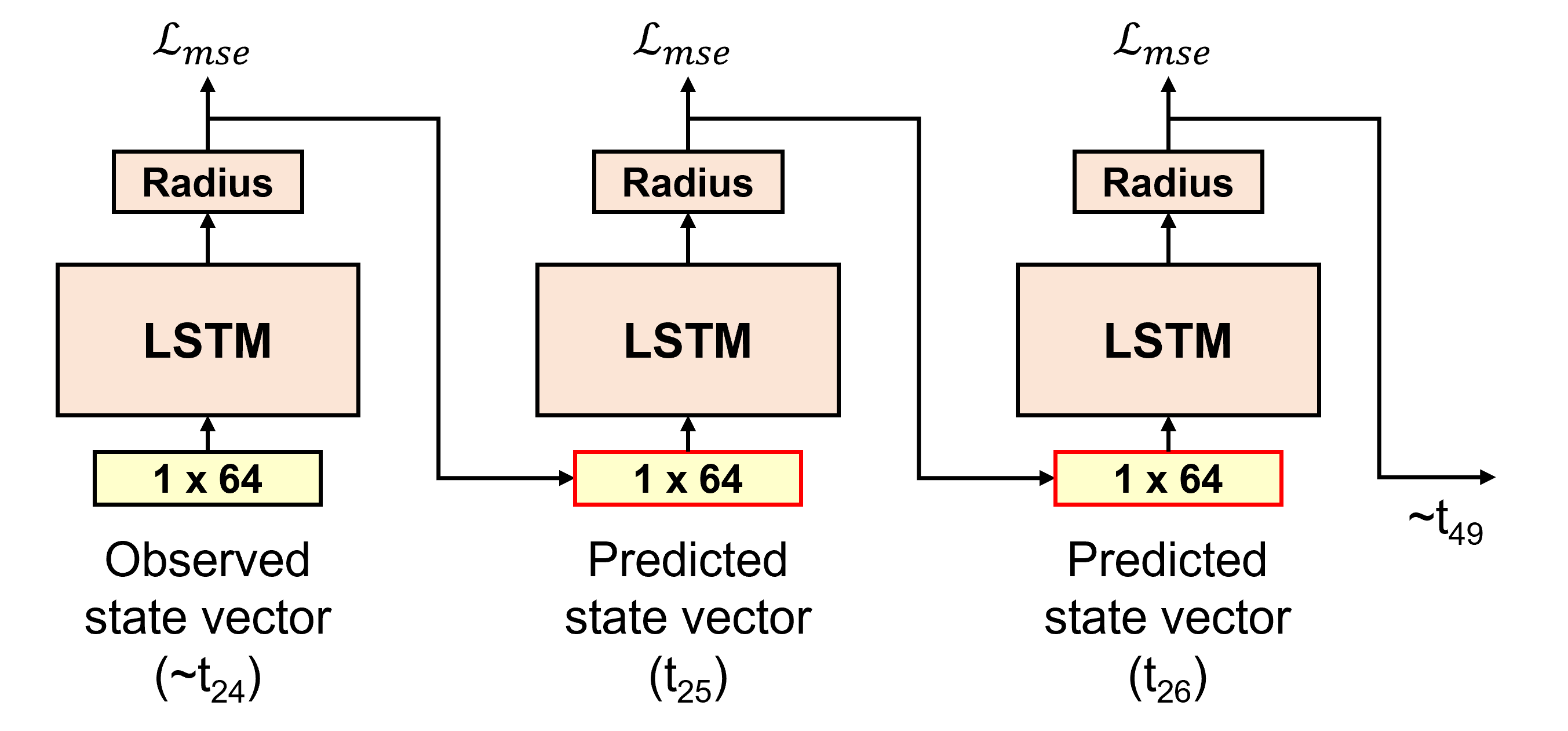}
\end{center}
\caption{Data flow in cluster prediction module. Output vectors in previous predictions are used for input to the next predictions.}
\label{figure_prediction_module}
\end{figure}

\subsection{Cluster Prediction Model}
Fig. \ref{figure_cluster_prediction_model} shows a schematic of our cluster prediction model. There are mainly two modules, the set-to-cluster module and prediction module. The figure describes how multiple agents are processed with the FC layer and MaxPooling layer. The set-to-cluster module transforms the positions of the observed agent sets to the state vector, which allows LSTM to learn temporal dynamics of the agent groups in fixed dimension space. Note, as we are learning and forecasting the cluster dynamics using the state vector, the predicted state vector can directly represent the future cluster configuration without other modules. The prediction module includes a single and unidirectional LSTM cell and two FC Layers at the input and output of the LSTM cell. It predicts the state vectors of the next time steps in an autoregressive way. LSTM learns temporal dynamics in the latent space, and a following FC layer predict cluster radii from the latent space. Another FC layer and LayerNorm layer are included between the set-to-cluster module and LSTM to normalize the distribution of elements in each state vector. We observe that the model performs better when LayerNorm is applied only in the observation stage.

\subsection{Training Cluster Prediction Model}
It is important to note that the set-to-cluster module is trained by unsupervised Hebbian learning that does not require a loss function. On the other hand, the prediction module is trained by supervised learning which ground truth should be prepared in advance. For the reason, the clustering module is first trained so that the ground truth is defined as output state vectors of the pre trained module. The weights of the set-to-cluster module is initialized by uniform distribution \(\mathcal{U}[0,1]\). As positions of the agents are normalized from 0 to 1, such initialization enables the more weight vectors to be effectively used as cluster centroids. Once the training is finished, cluster centroids are fixed and cluster radii are the only variable in the cluster representation. We save the state vectors of game frames before training the prediction module to reduce the running time of the clustering module.

Fig. \ref{figure_prediction_module} shows the data flow in the prediction module during repeated prediction. The module observes the first 25 state vectors, and then repeatedly predicts the next 25 state vectors by using previously predicted state vectors as the next input. Mean squre error (MSE) loss is used for the optimization, and the loss is calculated every time step. Hence, we optimize the loss from the total 49 state vectors. We set the learning rate 1e-3, batch size 16, and use Adam optimizer. The loss function is written by:
\begin{equation}
\mathcal{L}_{mse}(\textbf{s}, \hat{\textbf{s}}) =\frac{1}{N}\sum_{i}^{N}{\left(s_{i} - \alpha \hat{s_{i}} \right)^{2}}
\label{equation_mseloss}
\end{equation}

\noindent where N is the number of the elements in \(\textbf{s}\) and \(\textbf{y}\), and \(\alpha\) is an additional parameter to adjust the scale of cluster radii. While the map coordinate is normalized from 0 to 1, most of the cluster radii are smaller than 1e-1. It results in the linear approximation of MSE loss, decreasing the misprediction error for the large clusters. We heuristically set \(\alpha\) to be 10 to avoid the linear approximation.

\begin{figure}[t]
\begin{center}
\includegraphics[width=\columnwidth]{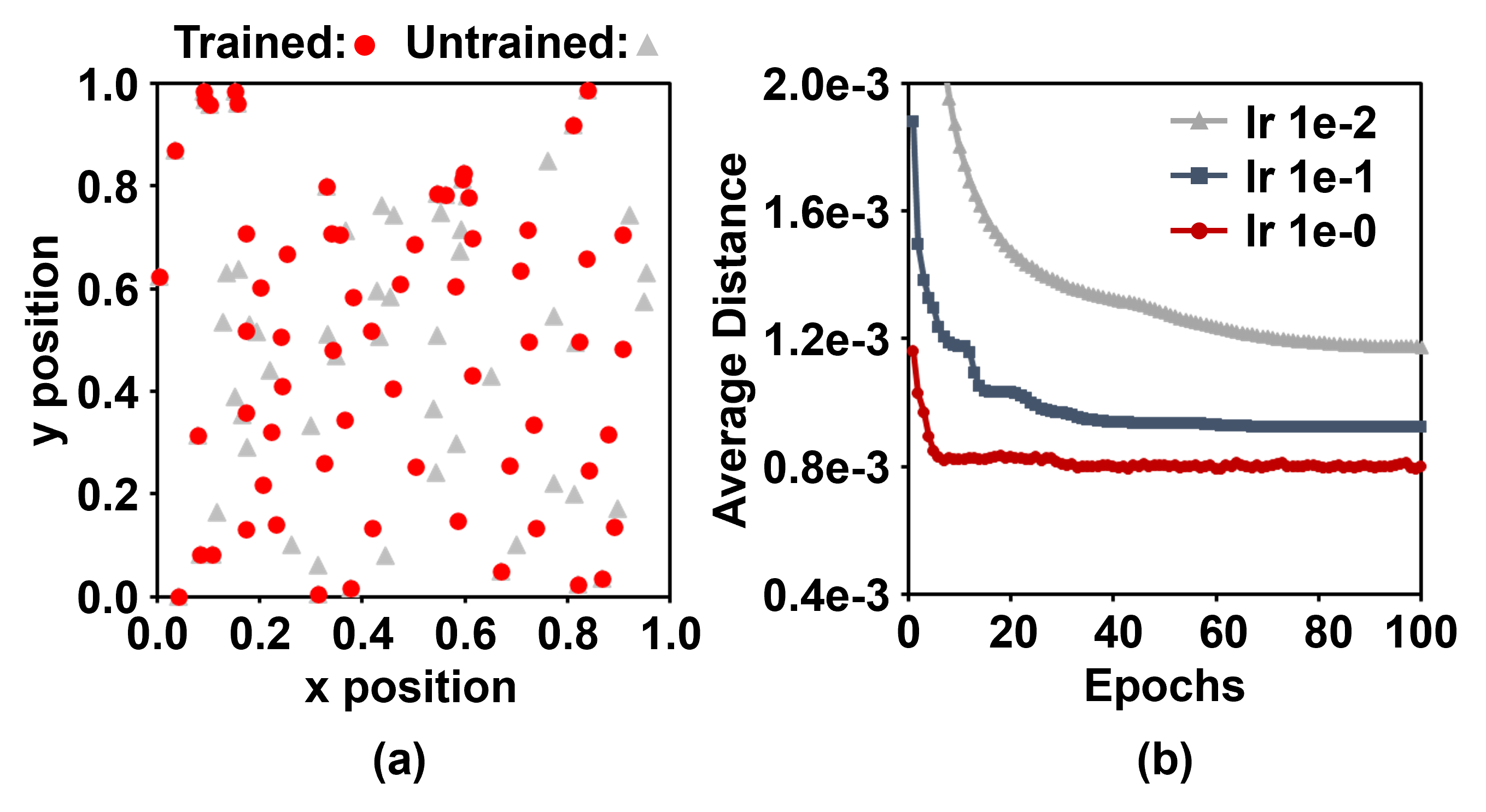}
\end{center}
\caption{Comparison between untrained and trained cluster centroids (a) and average distance between cluster centroids and agents during training (b).}
\label{figure_cluster_centroids}
\end{figure}

\section{Experimental Results}
\subsection{Dataset}
PySC2 is a Python-based machine learning environment for StarCraft II \cite{vinyals2017starcraft}. It supports to extract the useful properties of units in replay videos such as position and health. We are only interested in the position of units in this paper. Replay videos are uploaded on GitHub by Blizzard (https://github.com/Blizzard/s2client-proto). We randomly choose a Terran versus Terran game from the replay pack1 in the source. It has 11,000 frames that are captured every 0.2 seconds, and the resolution is set to (256, 256). We use 8,800 frames for training data, and 1,100 frames are used for validation and test data. For the temporal learning purpose, the frames are divided into 220 chunks where each chunk has successive 50 frames (i.e. 10 seconds). 176 chunks are used as training data, and rest two 22 chunks are used as validation and test data.

\begin{figure}
\begin{center}
\includegraphics[width=\columnwidth]{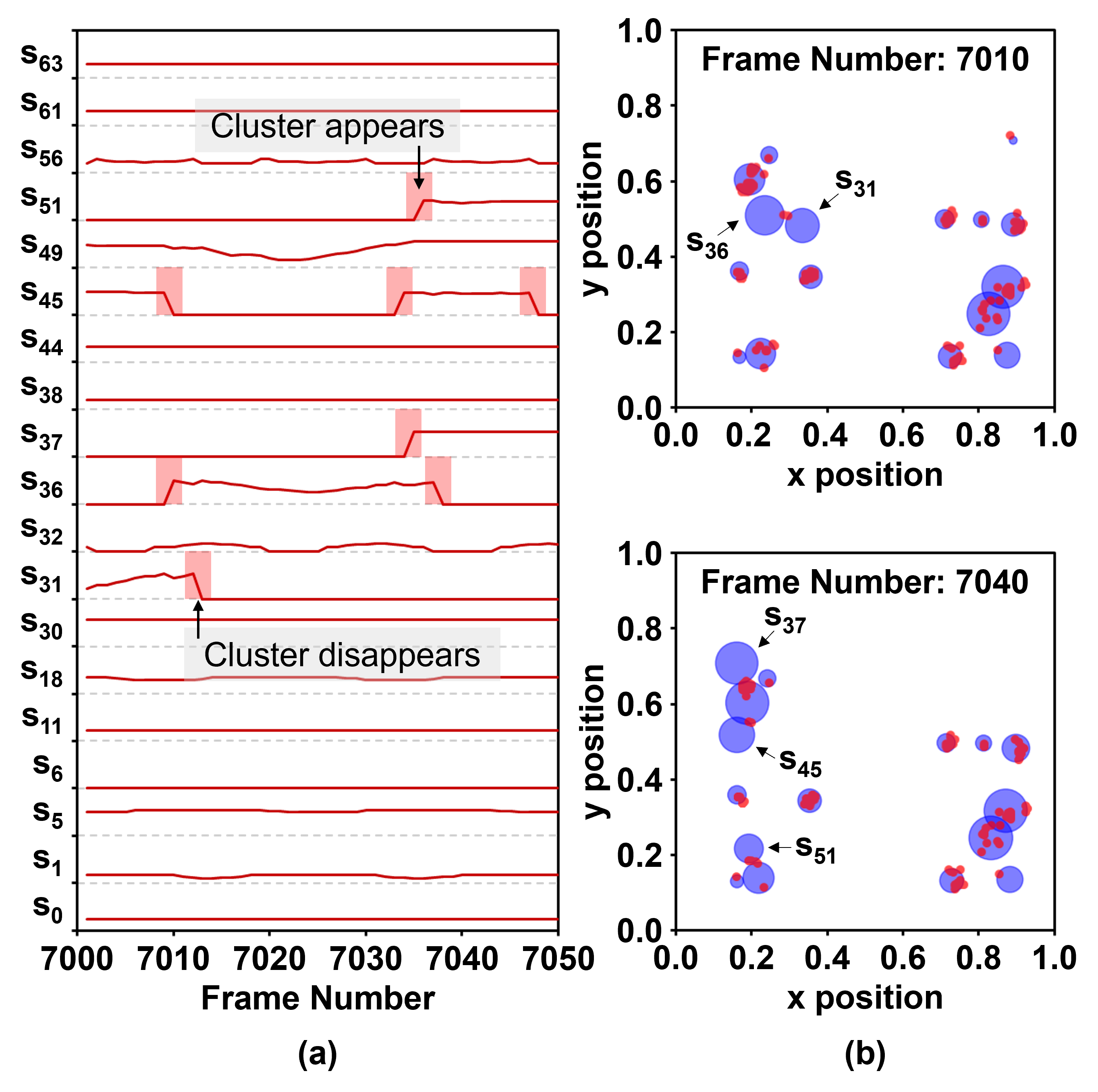}
\end{center}
\caption{State values in successive frames (a) and visualization of state values at certain frame (b).}
\label{figure_state}
\end{figure}

\subsection{Cluster Representation}
Fig. \ref{figure_cluster_centroids}(a) shows the cluster centroids before and after Hebbian learning. As our Hebbian learning updates weights to reduce the average Euclidean distance between the cluster centroids and agents, the trained centroids imply that the agents are more likely to be on them than the untrained centroids. The average distance is investigated to verify whether Hebbian learning actually reduces the distance. As the learning rule given in (\ref{equation_basic_rule}) is to minimize the distance, the cluster representation with smaller cluster radii is preferred. Fig. \ref{figure_cluster_centroids}(b) shows the distance during training epochs with different learning rates. Here, the distance (\(d\)) in a frame is defined by:
\begin{equation}
d=\frac{1}{N}\sum_{\textbf{x} \in S}{\underset{j}{\text{min}}{\|\textbf{x}-\textbf{w}_{j}\|_{2}}}
\label{equation_distance}
\end{equation}

\noindent where N is the number of agents and S is an agent set in a frame.  It means the average distance from agents to their closest cluster centroids. y-axis in Fig. \ref{figure_cluster_centroids}(b) is calculated by accumulating the distance (\(d\)) over all frames in validation data and dividing by the number of the frames. We observe the average distance 5.0e-3 in the untrained model while the trained models achieve minimum 0.8e-3. Hence, Hebbian learning decreases the average cluster radii required to represent the agent groups. However, there is no noticeable difference between trained and untrained centroids near edges and corners. We expect that these centroids are not often selected as winner neurons because they are too far from the frequently used paths of the agents. The learning rates higher than 1e-0 are not included in the figure as the average distance diverges in those settings. We set the learning rate 1e-0 and batch size 16 for the following experiments.

\begin{figure}[t]
\begin{center}
\includegraphics[width=\columnwidth]{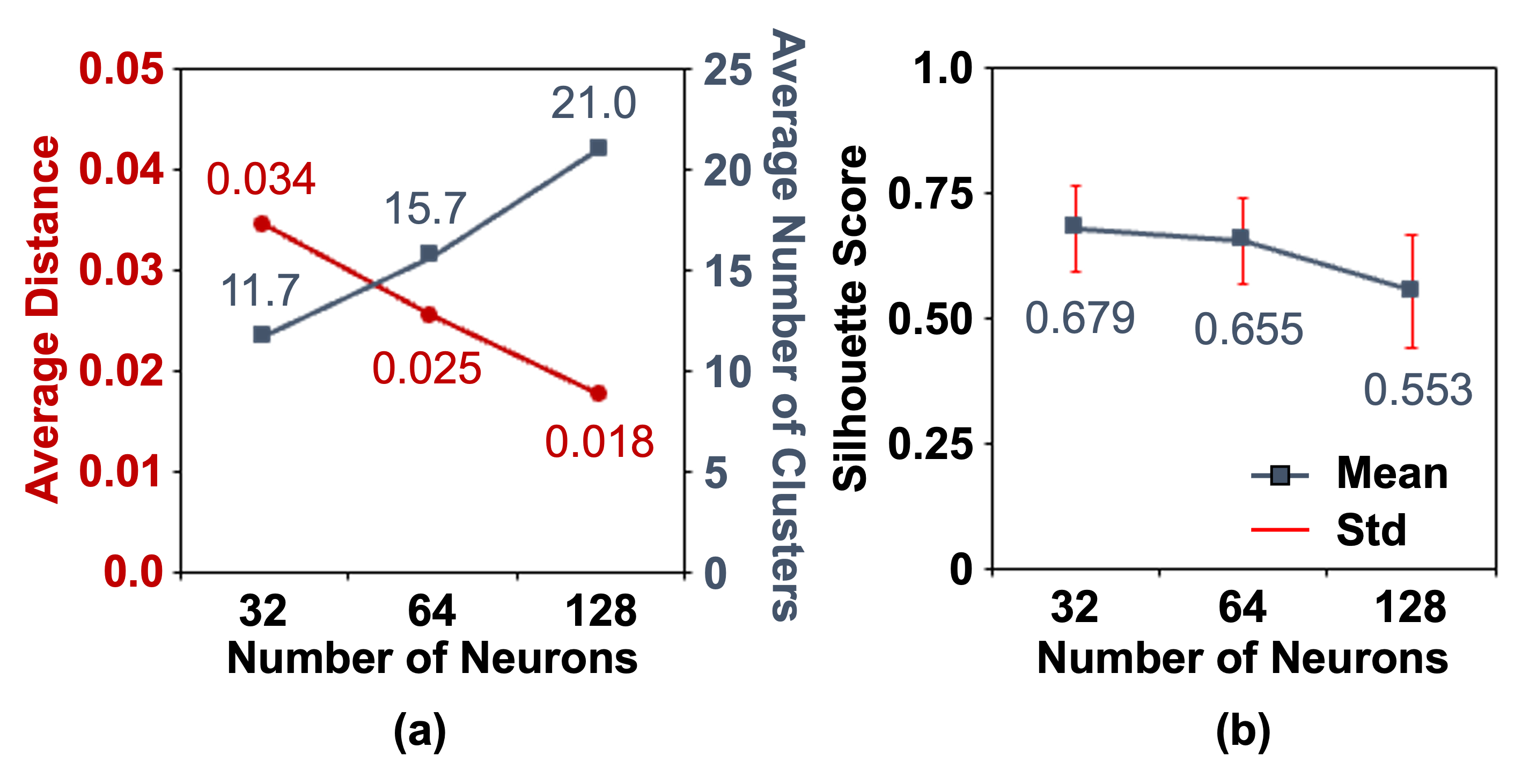}
\end{center}
\caption{Comparison of average distance between the agents and cluster centroids, average number of clusters (a), and Silhouette score depending on number of neurons (b).}
\label{figure_clustering_quality}
\end{figure}

Fig. \ref{figure_state}(a) is an example to show how the state vector temporally changes in successive frames. The y-axis ranges from 0.0 to 0.1 for each state. There are 19 neurons activated in the observed frames, so the other state values of loser neurons are omitted in the figure. The red boxes in the figure indicate sudden changes in the state values that imply the appearance or disappearance of clusters. As the state vector and weight vectors in the clustering module provide the information of cluster centroids and their radii, no additional decoding process is required. Fig. \ref{figure_state}(b) shows agent level (red) and cluster level (blue) representation at the frame number 7010 and 7040. There are initially two clusters near \((x,y)=(0.2,0.2)\) in the figure, and the new cluster appears above them in the later frame. It is because some agents in the initial cluster move upwards and the new cluster centroid is activated as they get closer to it. We observe the corresponding state value \(\textbf{s}_{51}\) clearly shows a sudden change during the two frame numbers in Fig. \ref{figure_state}(a). That is, the prediction module aims to forecast such sudden change in some states, which is resulted by the changes in the radius of neighbor clusters. This example only shows cluster division, but there happens similar scenarios in cluster merger as well.

\begin{figure}[t]
\begin{center}
\includegraphics[width=\columnwidth]{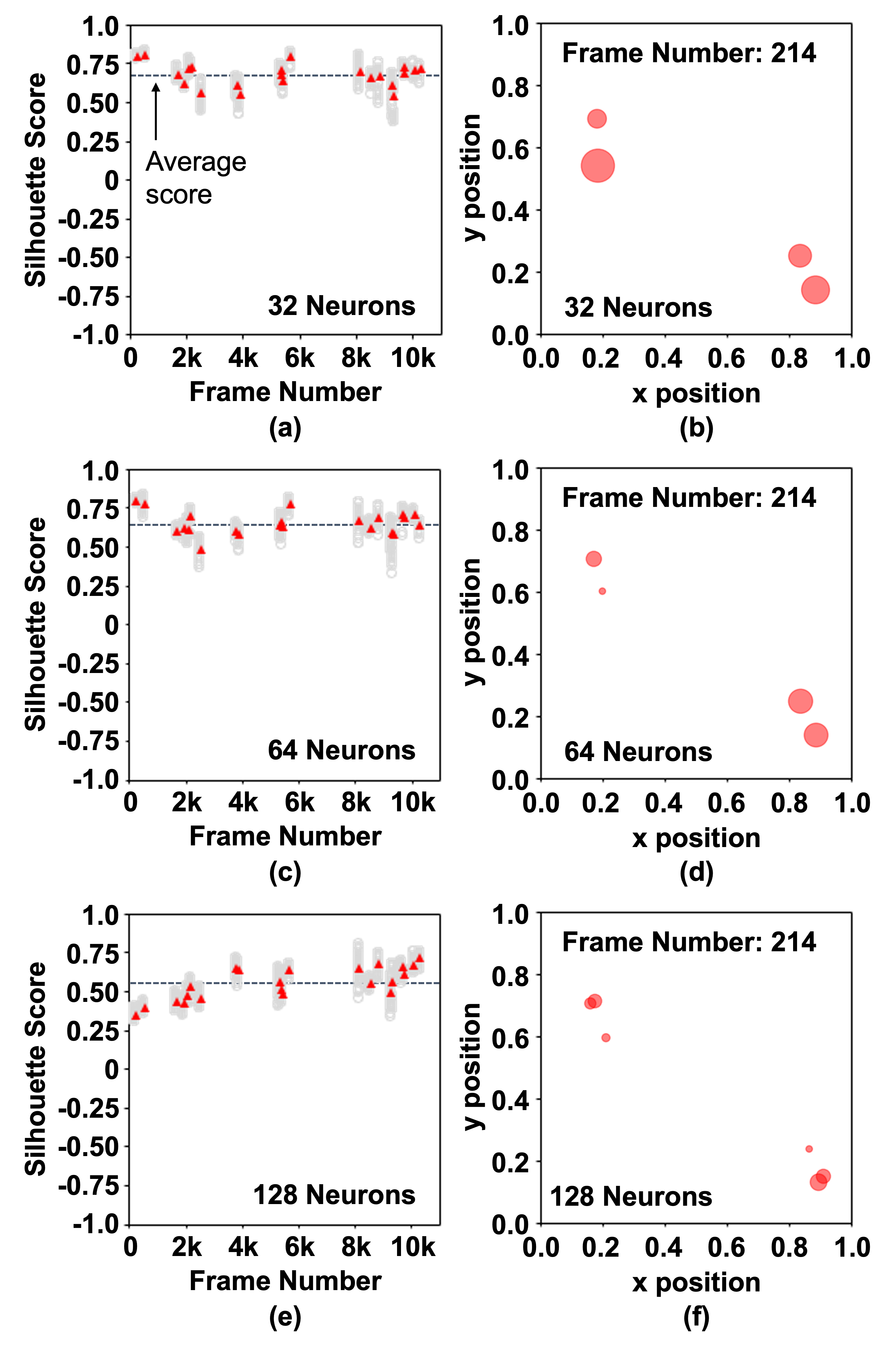}
\end{center}
\caption{Comparison of Silhouette score in each test frame depending on number of neurons (a, c, and e) and example of cluster representation of a test frame depending on number of neurons (b, d, and f).}
\label{figure_silhouette_score}
\end{figure}

\begin{figure*}[t]
\begin{center}
\includegraphics[width=\textwidth]{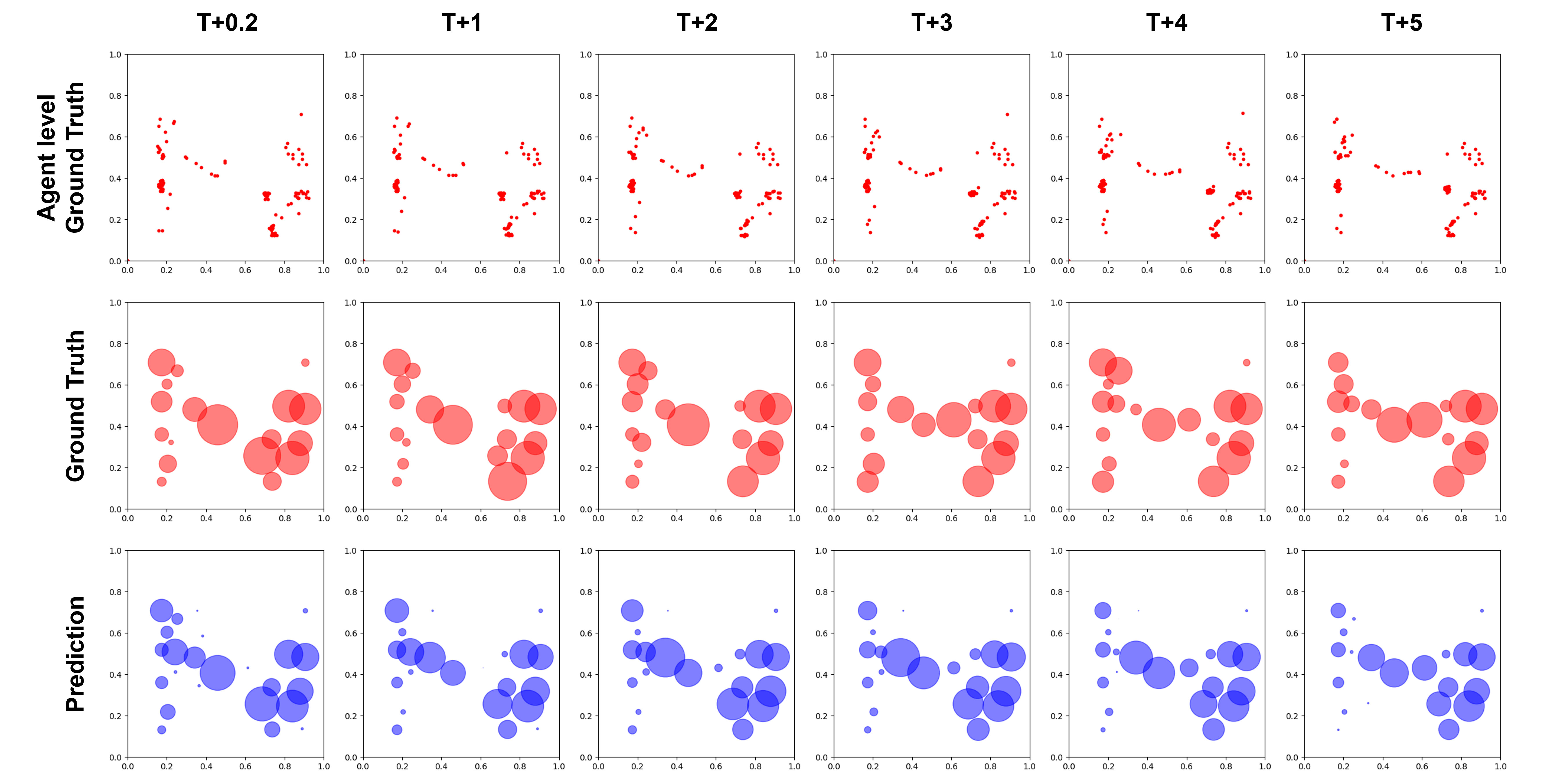}
\end{center}
\caption{Cluster prediction results with corresponding ground truth. Agent level ground truth is the original input to the model, and ground truth is the clustered one. 'T+0.2' and 'T+5' are the first and last frame in the prediction window.}
\label{figure_cluster_prediction}
\end{figure*}

\begin{figure}[t]
\begin{center}
\includegraphics[width=\columnwidth]{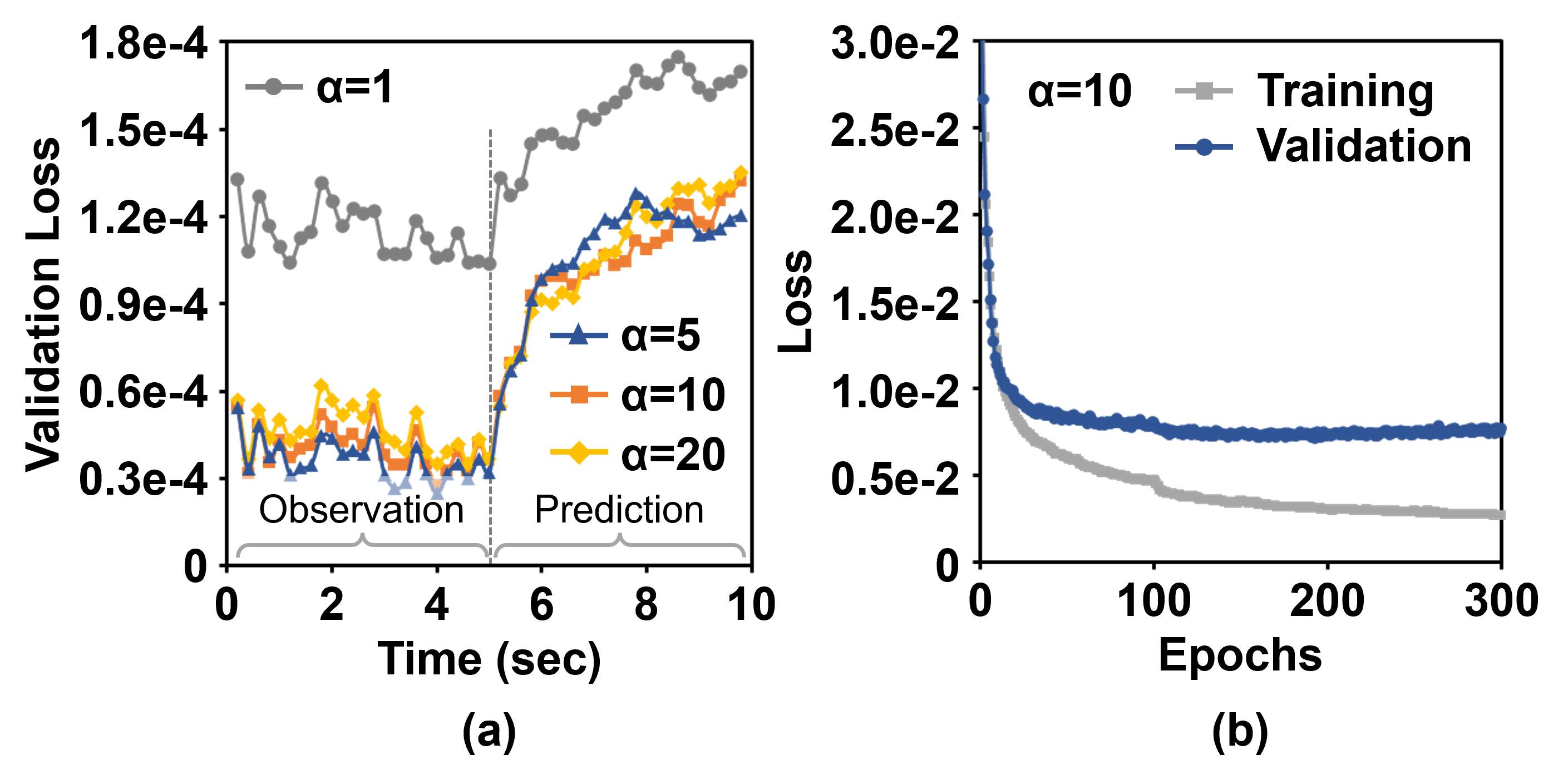}
\end{center}
\caption{Comparison of loss in observation and prediction window with different \(\alpha\) values (a) and loss changes with \(\alpha=10\) during training epochs (b).}
\label{figure_loss_alpha}
\end{figure}

\subsection{Clustering Quality}
A main control parameter in the set-to-cluster module is the number of output neurons that determines the maximum number of clusters. We study how the clustering quality varies depending on the number of the neurons. Fig. \ref{figure_clustering_quality}(a) shows the average Euclidean distance between the agents and associated cluster centroids, and the average number of created clusters in the test dataset. As the available cluster centroids are densely distributed in the module with the more neurons, the agents are likely to be grouped by the more centroids, thereby increasing the number of clusters and decreasing the distance. While more clusters generally allow smaller clusters which will increase the in-cluster similarity of clustered data points, it may cause the poor cluster-to-cluster variation by breaking the similar data points into different clusters.

Silhouette score is often used as a measure of clustering quality as it quantifies both the in-cluster similarity and out-cluster dissimilarity \cite{rousseeuw1987silhouettes}. It is mathematically described by:
\begin{equation}
\label{eq_silhouette1}
s(i)= \frac{b(i) - a(i)}{\text{max}(a(i), b(i))}
\end{equation}

\begin{equation}
\label{eq_silhouette2}
a(i)= \frac{1}{|C_{I}|-1} \sum_{j \neq i, j \in C_{I}}{\|\textbf{x}_{i}-\textbf{x}_{j}\|_{2}}
\end{equation}

\begin{equation}
\label{eq_silhouette3}
b(i)= \underset{J \neq I}{\text{min}}(\frac{1}{|C_{J}|} \sum_{j \in C_{J}}{\|\textbf{x}_{i}-\textbf{x}_{j}\|_{2}})
\end{equation}

\noindent where \(s(i)\) is the Silhouette score of the \(i\)-th data point. Note, the score ranges from -1 to 1, and the higher score is preferred. \(|C_{I}|\) indicates the number of data points in the \(I\)-th cluster. \(a(i)\) and \(b(i)\) are the average in-cluster distance and out-cluster distance of the \(i\)-th data point, respectively. Fig. \ref{figure_clustering_quality}(b) shows the mean and standard deviation of Silhouette score, where all the agents are considered in the test dataset, with a variable number of the neurons. The mean and standard deviation of the score at the 128 neurons is lower and higher than the others while the 32 and 64 neurons achieve the similar quality.

The clustering quality with the 128 neurons is particularly poor at the early stage of the game, where only few clusters are enough to represent the agents. Fig. \ref{figure_silhouette_score} shows Silhouette score in each test chunk. In Fig. \ref{figure_silhouette_score}(a), (c), and (e), the x value of data points represents the average frame number of their chunk. The grey data points at the x value indicates Silhouette scores of the 50 frames in the chunk, and the red data point is the average of them. There are 4 clusters in Fig. \ref{figure_silhouette_score}(b) and (d), which is reasonable as there are the few agents in the game. However, in Fig. \ref{figure_silhouette_score}(f), 6 clusters are unnecessarily created though they are smaller than the clusters created by the 32 neurons and 64 neurons. It then decreases the out-cluster distance in (\ref{eq_silhouette3}) and the score as shown in Fig. \ref{figure_silhouette_score}(e). Hence, less than the 64 neurons will be enough to create a variable number of clusters in the game without sacrificing the clustering quality. Our other experimental results are based on the module with the 64 neurons.

\subsection{Cluster Prediction}
Our main task is to forecast complex evolution of the clusters. Fig. \ref{figure_cluster_prediction} displays predicted configurations of the clusters and corresponding ground truth in a chunk. While the model is not trained by looking at individual agent position, agent level representations are also incorporated in the figure to illuminate how the agents are clustered. As the duration of each chunk data is 10 seconds, the predicted results are made for 5 seconds after observing the movement of the clusters for the first 5 seconds. 

We observe our prediction module is able to forecast complex evolution of the clusters during the prediction window. For example, the agent level ground truth shows an interesting movement of the agents. Few agents near the center area at 'T+0.2' are moving toward the right bottom direction. However, they start to horizontally move, connecting the left and right areas in the map at 'T+5'. As we assume the radial cluster shape, such horizontal arrangement of the agents are represented with large clusters in the cluster level ground truth. The predicted clusters show that the corresponding cluster centroids are successfully activated at 'T+5'. Also, most of the clusters at other areas are reasonably predicted though few clusters are not activated or remained. The predicted results are made after training the model for 300 epochs with the learning rate 1e-3 and batch size 16.

Fig. \ref{figure_loss_alpha}(a) shows the average validation loss in each time step with different \(\alpha\) values. The models are trained in the same configuration except \(\alpha\) in the loss function. However, the loss in the figure is calculated without the \(\alpha\) so that the values are compared in the same metric. We observe that the model performance clearly degrades without the consideration of \(\alpha\) (i.e. \(\alpha=1\)) while the validation loss is similar in other values. Given the maximum radius of the clusters in the training data is 0.112, negligible changes in cluster radii would make learning dynamics of the clusters difficult. Our experimental results are based on the model trained with \(\alpha=10\). Fig. \ref{figure_loss_alpha}(b) shows the training and validation loss for 300 epochs in \(\alpha=10\) setting, and the model starts to overfit to the training data after 100 epochs.

\subsection{Failure case}
\begin{figure}[t]
\begin{center}
\includegraphics[width=\columnwidth]{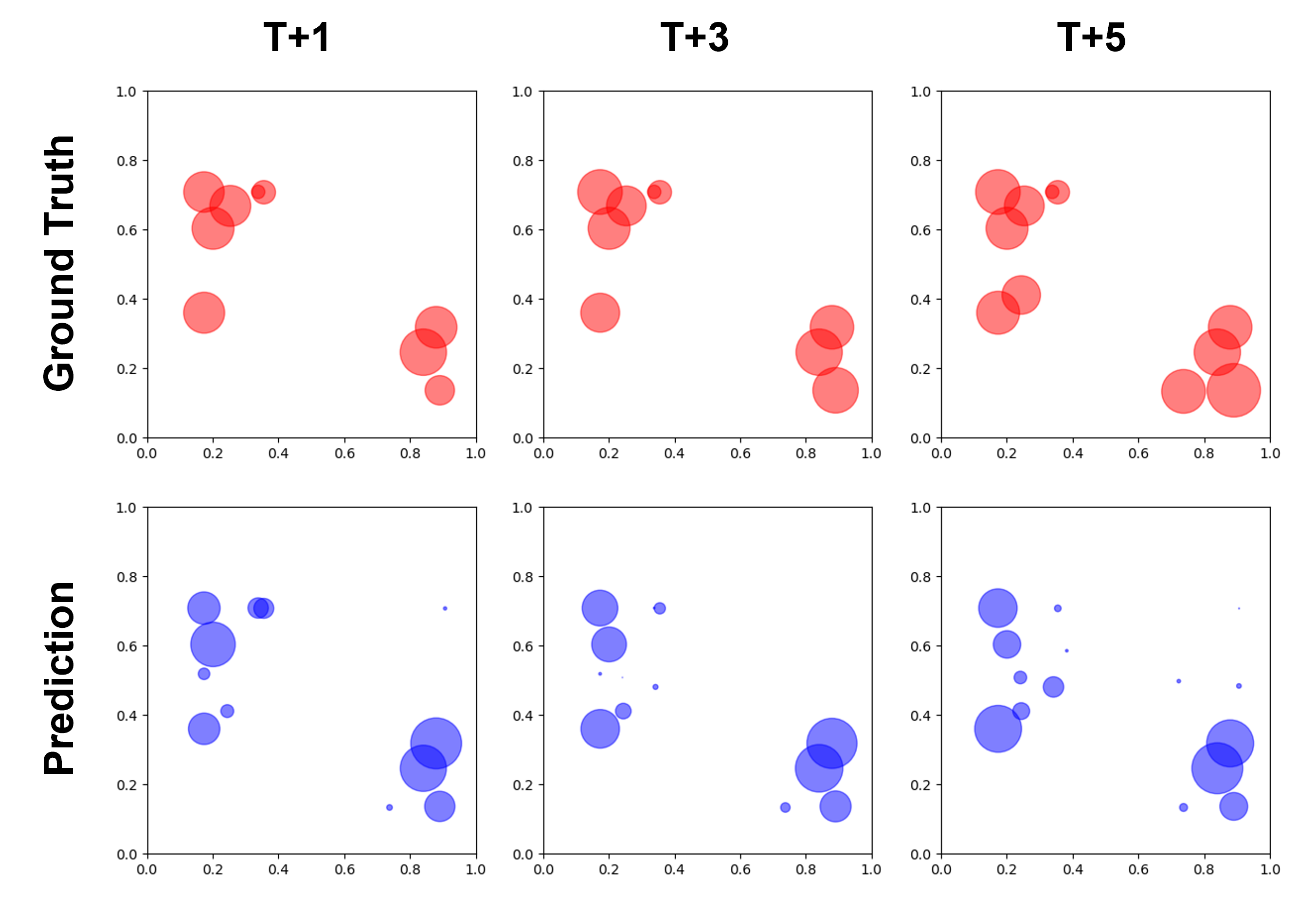}
\end{center}
\caption{Cluster prediction results with corresponding ground truth in a failure case. Most of the clusters in the ground truth have no active changes during the prediction window.}
\label{figure_failure_analysis}
\end{figure}

We observe the reasonable prediction of the dynamically moving clusters in Fig. \ref{figure_cluster_prediction}. However, there also exist few clusters failed to be predicted. Fig. \ref{figure_failure_analysis} shows an example in which the model fails to correctly predict the newly appeared clusters in the early stage of the game. Given active battles are not likely to be happened early, most of the clusters are not actively moving, rather remain to motionless. For example, in the ground truth, there is the one cluster at the left middle area until 'T+4', and the new cluster suddenly appears at 'T+5'. As there is no noticeable change in the single cluster with regard to its radius before 'T+4', it seems the agent in the cluster suddenly starts to move creating the new cluster. Such discontinuous and uncertain dynamics in the clusters are failed to be predicted by the model. However, the figure shows that the new cluster is predicted in advance with a small radius, and similar result is also observed in the right bottom area. While the radii are not accurate, the model makes the prediction with the non zero radii at the probable cluster centroids.

\subsection{Comparison with Other Methods}
As it is not available to directly compare with other prediction models due to no benchmarking, we mainly compare the proposed clustering module with other clustering methods. Our clustering method is similar with K-means clustering in terms of minimizing the Euclidean distance \(\|\textbf{x}-\textbf{w}\|_{2}\). A difference in our method is lower inference time complexity. As there is no difference between training and inference in k-means clustering, the time complexity of approximated k-means clustering is given by \(\mathcal{O}(n \times k \times i)\) where \(n\) is the number of points, \(k\) is the number of clusters, and \(i\) is the number of iterations. Our method also has the same time complexity \(\mathcal{O}(n \times k \times i)\) for training, but it becomes \(\mathcal{O}(n \times k)\) for inference. It is because the FC layer in the set-to-cluster module stores the cluster centroids as the weight vectors in advance, hence we do not need to repeatedly optimize the centroids again during the inference. 

Also, the set-to-cluster module enables to create a variable number of clusters while K-means clustering creates K clusters regardless of the configuration in the agents. However, K-means clustering is a non-parametric method while our method requires the FC layer to store the pre-trained centroids. Table \ref{table_comp} shows the computational complexity of the modules per prediction for a frame. The single FC layer does not incorporate the large number of weight parameters, so the overhead of the clustering module is negligible compared to the parameters of the prediction module.

Hierarchical clustering and DBSCAN (Density-based spatial clustering of applications with noise) are the widely used clustering methods that are able to create a variable number of clusters. However, hierarchical clustering needs to specify the number of clusters at the end, so it does not automatically choose how many clusters are required. While DBSCAN can represent the variable number of clusters based on the spatial density of data points (i.e. game agents), it still requires repeated how to represent the cluster configuration into a  irregular shape of the clusters will be hard to be represented in a simple state vector.

\begin{table}
\centering
\caption {Computational complexity of cluster prediction model}
\begin{tabular}{ccccc}
\hline
\\ [-0.8em]
\textbf{Model} & \textbf{Parameters} & \textbf{Activations} & \textbf{FLOPs} \\
\\ [-1em]
\hline
\\ [-0.8em]
Set-to-cluster module & 128 & 6.5k & 25.9k \\
\\ [-1em]
Prediction module & 41.7k & 704 & 81.9k \\
\\ [-1em]
\hline
\end{tabular}
\label{table_comp}
\end{table}

\section{Conclusions}
This paper proposes a Hebbian learning based spatial clustering method for game agents in StarCraft II, along with a prediction module to forecast the complex evolution of the clusters. Our method efficiently deals with a variable number of clusters and shows lower time complexity compared to K-means clustering. With the cluster representations, the prediction module learns the dynamics of the clusters and predict the cluster radii in the future. Extending its application to other multi-agent systems where agents frequently create groups would be an interesting direction for future research.

\bibliographystyle{IEEEtran}

\bibliography{Reference}

\end{document}